\documentclass[10pt,twocolumn,letterpaper]{article}

\usepackage{multirow}
\usepackage{booktabs}
\usepackage{placeins}
\usepackage{wacv}              

\definecolor{wacvblue}{rgb}{0.21,0.49,0.74}
\usepackage[pagebackref,breaklinks,colorlinks,allcolors=wacvblue]{hyperref}

\def\wacvPaperID{846} 
\def\confName{WACV}
\def\confYear{2027}

\title{What Makes Adversarial Examples Transfer Across Deepfake Detectors? }

\author{
Rafael M. Mamede$^{1,2}$ \qquad
Pedro C. Neto$^{2,3}$ \qquad
Ana F. Sequeira$^{1,2}$\\
$^{1}$INESC TEC, Porto, Portugal\\
$^{2}$Faculty of Engineering, University of Porto (FEUP), Porto, Portugal\\
$^{3}$Unilabs, Porto, Portugal\\
{\tt\small
\{rafael.c.maia,ana.f.sequeira\}@inesctec.pt}\\
{\tt\small pedro.neto@unilabs.com}
}

\begin{document}
\maketitle
\begin{abstract}
Deepfake detectors remain vulnerable to transfer-based black-box attacks, in which adversarial examples are generated on a source surrogate model and transferred to a target model, unknown to the attacker. Yet how source--target compatibility shapes attack success remains poorly understood. Prior studies evaluate limited detector pools and rarely disentangle architectural from training factors. We conduct a controlled evaluation of adversarial transferability across 60 detectors spanning six backbones, two pretraining regimes, and five training-data configurations, using two attack procedures: AutoAttack (AA) and the Carlini--Wagner attack with Expectation over Transformation (CW--EOT). Matched comparisons reveal significantly higher transfer when source and target share an exact backbone, architecture family, pretraining regime, or training data. This compatibility structure is attack-dependent: exact backbone compatibility has the largest effect under AA, whereas shared pretraining and training data have the largest effects under CW--EOT. When transfer is averaged across non-target sources, mean attack success rate (ASR) is $7.21\%$ under AA and $19.52\%$ under CW--EOT. By contrast, a multi-source oracle combining both attacks attains a \(64.48\%\) mean ASR after excluding exact backbone and training-data matches, showing that source averaging can substantially understate target vulnerability. We release 240,000 adversarially perturbed images, complete pairwise transfer results, detector configurations, and evaluation code. These findings establish source--target compatibility and source-model selection as central dimensions of credible transfer-based black-box robustness evaluation.

\end{abstract}

\section{Introduction}

As generative models become more widespread, so too becomes their potential for misuse. Images and videos manipulated or generated using artificial intelligence (AI) have become harder to distinguish from real content, contributing to growing distrust in media \cite{deepfakes_and_disinformation}. This, in turn, has motivated the development and evaluation of deep learning based detectors of AI-generated content \cite{roessler2019faceforensicspp,DeepfakeBench_YAN_NEURIPS2023,yan2024df40,DFDC2020,Celeb_DF_cvpr20}.

\begin{figure}[!t]
\centering
\includegraphics[width=\columnwidth]{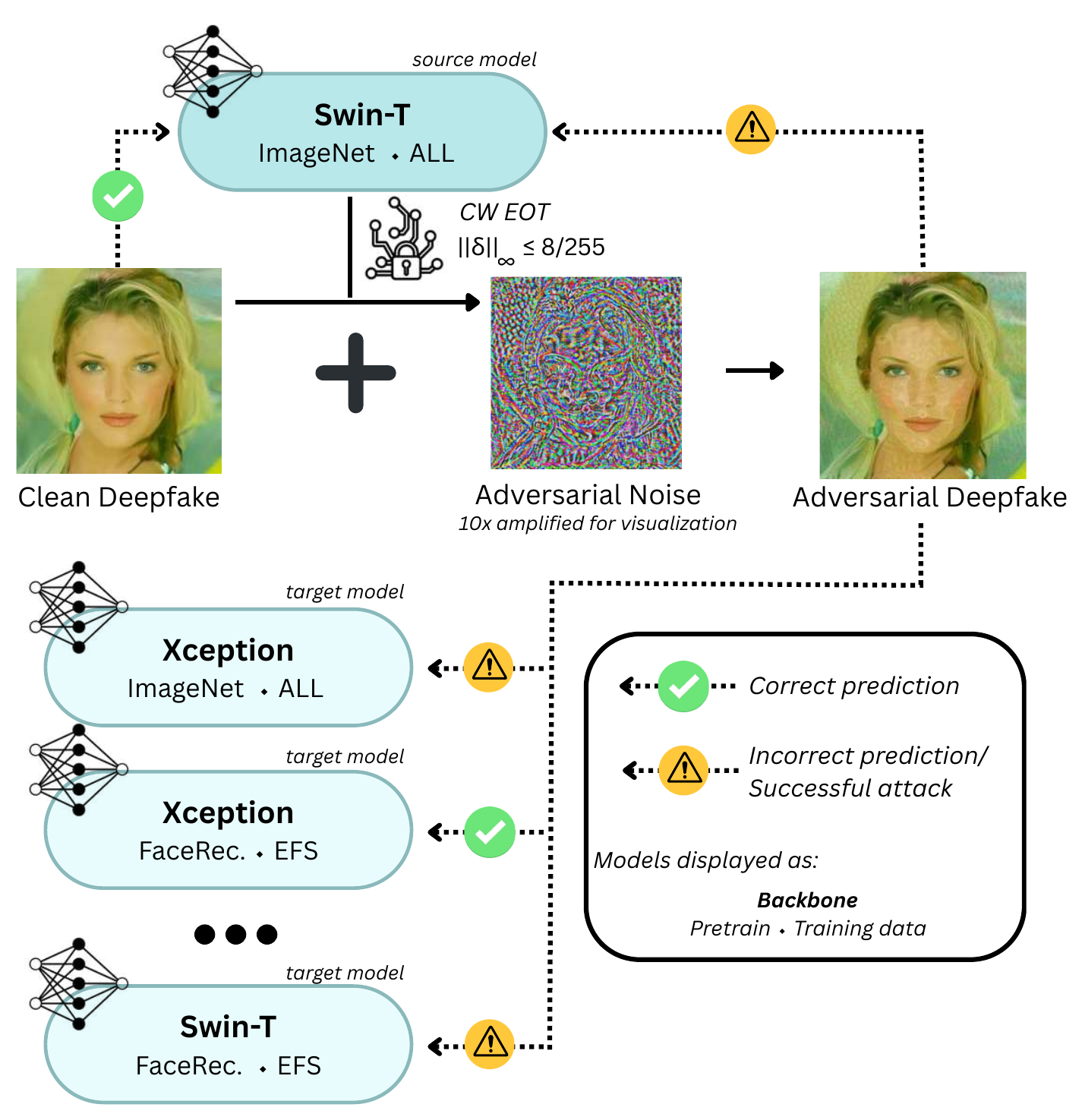}
\caption{Overview of the transfer evaluation. An $\ell_\infty$-bounded perturbation ($\epsilon=8/255$) is generated on a source detector and evaluated on targets differing in backbone, pretraining, and training subset.}
\label{fig:intro_fig}
\end{figure}

For more than a decade, we have known that neural networks are prone to being misled by adversarial examples \cite{szegedy2013intriguing,goodfellow2015explaining}. These carefully crafted manipulations consist of small perturbations in the image space, often imperceptible to humans, that change the model's decision. While initially, attacks were crafted using only perfect information of the model, as the field matured attacks have become more realistic. One key discovery was that adversarial examples can transfer between models, so an attacker can generate perturbations on a surrogate model to try to mislead a different unknown model \cite{szegedy2013intriguing}.

On the other hand, the detection of Deepfake content (AI-manipulated or fully generated content depicting real or fictional people) has been shown to suffer from generalization problems detecting images produced by unseen generators \cite{yan2024df40}. This suggests that detectors may learn cues that are too specific to the data and generation methods seen during training, rather than generalized evidence of manipulation. Thus, it remains unclear whether adversarial examples crafted on one detector exploit model-specific weaknesses, or whether they capture broader vulnerabilities that persist across different training conditions and unseen types of generated content.

In this work, we perform an in-depth study of adversarial transferability in Deepfake detection. We investigate how adversarial examples generated on surrogate detectors transfer to target detectors trained with different architectures and different sets of generation methods. This allows us to examine whether transferability is mainly driven by shared model architecture, by similarities in the training and pretraining data, or by broader vulnerabilities common to Deepfake detectors. 
Thus, our \textbf{main contributions} are as follows:
\begin{itemize}
\item We present a controlled, large-scale evaluation of black-box adversarial transfer across Deepfake detectors. We construct a factorial bank of 60 detectors spanning six backbones from convolutional neural network (CNN) and Transformer architecture families, two pretraining regimes, and five training-data subsets, producing 3,540 ordered black-box source--target pairs per attack.

\item Through matched source--target comparisons, we show that transfer is systematically associated with exact backbone, architecture family, pretraining regime, and training-data compatibility. This structure is attack-dependent: exact-backbone compatibility has the largest effect under AA (\(22.34\) pp), whereas pretraining-regime and training-data compatibility have the largest effects under CW--EOT (\(21.25\) and \(19.03\) pp, respectively).

\item We show that single-source and source-averaged transfer evaluations can substantially understate target vulnerability. We complement these evaluations with multi-source oracle protocols that provide empirical upper-bound vulnerability over eligible source pools. Even when sources using the target's exact backbone or containing its complete manipulated-image training set are excluded, the combined strict oracle reaches a mean ASR of \(64.48\%\).

\item We release a reproducible benchmark comprising 240,000 adversarially perturbed images, complete pairwise transfer results, detector configurations, and evaluation code, supporting compatibility-aware evaluation of black-box robustness\footnote{Code and benchmark resources are available at \url{https://github.com/RafaelMMamede/deepfake-adversarial-transfer}}.

\end{itemize}
\section{Related Work}

In this section, we review the evolution of Deepfake generation and detection, before discussing adversarial attacks against Deepfake detectors and the transfer of adversarial examples across models.

\textbf{Deepfake Generation and Benchmark Diversity.}
Early Deepfake benchmarks predominantly represented a limited set of facial manipulation mechanisms. FaceForensics++ brought together four widely used manipulation methods, including identity swapping, facial reenactment, and neural rendering, while Celeb-DF and DFDC increased the realism and scale of face-swapping data \cite{roessler2019faceforensicspp,Celeb_DF_cvpr20,DFDC2020}. More recent benchmarks have expanded the scope of Deepfake detection beyond these settings. In particular, DF40 collects 40 manipulation techniques and organizes them into four broad families: face swapping (FS), face reenactment (FR), entire-face synthesis (EFS), and face editing (FE)
\cite{yan2024df40}. These families produce substantially different visual traces and artifact distributions. Consequently, detectors trained on a restricted set of manipulation techniques may learn method-specific cues that do not persist nor generalize across other forms of generated or edited content.

\textbf{Deepfake Detection and Generalization.}
Deepfake detection is commonly formulated as binary classification between authentic and manipulated facial content. Most established approaches rely either on convolutional neural networks operating in the spatial domain or on transformer-based visual encoders. Other approaches exploit frequency-domain cues, while video-based detectors may additionally model temporal inconsistencies across frames. DeepfakeBench provides a unified implementation and evaluation framework, highlighting the considerable methodological diversity of the field \cite{DeepfakeBench_YAN_NEURIPS2023}.

One recurring issue is that strong performance reported under in-distribution evaluation does not necessarily persist when the training and evaluation distributions differ. Therefore, generalization performance is commonly assessed through cross-dataset and cross-manipulation evaluations, in which detectors are tested on datasets or forgery techniques not observed during training. These evaluations have repeatedly revealed substantial performance degradation under domain shifts \cite{DeepfakeBench_YAN_NEURIPS2023,electronics13030585,yan2024df40}.

Although generalization performance is often studied under benign distribution shifts, comparatively little attention has been given to whether the same detector and training factors also influence adversarial transferability.

\textbf{Adversarial Vulnerabilities.}
As most modern Deepfake detectors are based on deep neural networks, it is not surprising that they inherit the adversarial weaknesses that have been broadly identified in the field. Early studies confirmed this concern, where literature models were shown vulnerable to adversarial perturbations both on the image space, as well as the latent space of the generator \cite{carlini2020evading}. Furthermore, both white-box (assuming full model access from the attacker), and black-box (assuming no or little model access) attacks have been shown to remain effective after common image and video processing operations, including compression \cite{Neekhara2020AdversarialDE}. AVA further demonstrated that detectors can be bypassed through semantics-preserving changes to facial attributes in the generator's latent space \cite{AVA}.

More directly related to our work, Neekhara \textit{et al.} studied adversarial transfer across four Deepfake detection pipelines derived from leading DFDC submissions \cite{DFDC2020,Neekhara2021PracticalPerspective}. Using Expectation over Transformation (EoT), they crafted adversarial examples robust to translation, resizing, and additive noise to account for differences in face detection and preprocessing across pipelines. They also adapted universal adversarial perturbations to Deepfake video detection, showing that a single perturbation could be reused across frames and videos and transferred to unseen pipelines. However, their evaluation was limited to four CNN-based pipelines from three submissions. Because classifier architecture, face extraction, preprocessing, and augmentation varied jointly, the study could not isolate which source--target properties explained the observed variation in transferability.

More recently, Serrano \textit{et al.} adapted the DUMB/DUMBer framework to evaluate Deepfake detectors under realistic attacker--defender mismatches \cite{Serrano2026DUMB}. Their study considered five detectors, three attacks, and two datasets, organizing the evaluation into white-box, cross-model, cross-dataset, and combined cross-model and cross-dataset scenarios. Within this framework, they measured the effectiveness of transferred attacks and assessed whether adversarial-training strategies retained their benefits under each mismatch. Their results showed that attacks remained effective under model and dataset mismatch, whereas the benefits of adversarial training became less reliable under cross-dataset evaluation. However, transfer was summarized within broad model and dataset mismatch scenarios, consistent with their objective of evaluating attack and defense performance. Consequently, their analysis shows that transfer can persist under mismatch, but does not determine which source--target compatibilities make it stronger or whether the dominant compatibilities change across attacks.

In sum, these studies establish adversarial transfer as a practical threat to Deepfake detectors. However, existing analyses of adversarial example transferability have either considered a limited set of complete detection pipelines or organized transfer through broad model and dataset mismatch categories, primarily to improve transferable attacks or assess adversarial defenses. Our work treats variation in transferability as the primary object of study. We construct a systematically varied detector bank and examine how backbone, architectural family (CNN-based or Transformer-based), pretraining, and training data are associated with transfer across source--target detector pairs. With this methodology, we complement prior work by moving beyond demonstrations of whether transfer occurs toward a controlled, factor-based characterization of the conditions under which it becomes stronger or weaker.
\section{Methodology}

This work seeks to systematically characterize adversarial transferability across Deepfake detectors. To do so, we consider three main factors that we control for our detectors: pretraining, training data, and model architecture. We constructed a factorial
bank of detectors by combining six backbones, two pretraining strategies, and five training subsets.  

The selected backbones consist of three CNNs (ResNet34 \cite{resnet}, Xception \cite{xception}, and EfficientNet-B4 \cite{efficient}) and three Transformers (DeiT-S \cite{deit}, ViT-B/16 \cite{vit}, and Swin-T \cite{swin}). We also consider pretraining on either ImageNet \cite{imgnet}, or Face Recognition (on the BUPT-BalancedFace dataset \cite{balancedface}, with ElasticArcFace+ loss \cite{elasticface}, and validation on RFW \cite{rfw}). The detectors are then trained on subsets of DF40 \cite{yan2024df40}, with images sourced from FF++, and different types of manipulations: face-swapping (FS), face-reenactment (FR), entire-face-synthesis (EFS),  face-editing (FE), and a union of all manipulations (ALL).

This resulted in a model bank,  $\mathcal{M}$, with
\(
|\mathcal{M}| = 6 \times 2 \times 5 = 60
\)
trained detectors.

We then construct an attack image subset of 2,000 manipulated test images, and perturb them using two evasion attack methods for each source model in our detector bank, resulting in \(2{,}000 \times 60 \times 2 = 240{,}000\) perturbed images.

We treat transfer as directional, so each ordered pair of source and target models \((s,t) \in \mathcal{M} \times \mathcal{M}\) constituted a distinct evaluation. For each attack, this produced $60^2=3{,}600$
source--target evaluations, comprising 60 white-box (\(s=t \)) and 3,540 black-box pairs (\(s\neq t \)).

The experimental analysis comprises three components:
\begin{enumerate}
    \item \textit{Clean-performance assessment}. We first verify that the trained detectors retain utility on their intended task by measuring AUC (area under the receiver operating characteristic curve) on three clean (unperturbed) DF40 test sets. Two test sets contain samples from all manipulation families and differ in the underlying real-image dataset, using FF++ and CDF, respectively. The third test set contains only the manipulation family used to train the evaluated detector (and images from FF++), providing a matched in-domain evaluation.

    \item \textit{Adversarial risk assessment}. We assess each detector using white-box attacks and two black-box oracle settings. The \textbf{self-excluding oracle} selects the strongest attack for each image from all eligible surrogate models, estimating risk when the attacker has access to a diverse surrogate bank. The \textbf{strict oracle} excludes surrogates sharing the target's backbone or training data, measuring vulnerability under limited model similarity.

    \item \textit{Adversarial transferability study}. We evaluate each adversarial image against every target detector and measure transferability for all directional black-box source--target pairs. We then analyze how transfer is influenced by shared backbone, architecture family, pretraining strategy, and deepfake data used for training, using matched comparisons and statistical hypothesis tests.

\end{enumerate}

All detectors are trained for binary real-versus-manipulated classification using the FaceForensics++ based partition of DF40. The operational threshold was chosen based on maximizing balanced accuracy in a held-out validation set with multiple manipulations sampled from the FaceForensics++ based partition of DF40. We consider a fixed set of 2,000 manipulated test images to be attacked, balanced across the four manipulation families and spanning 36 manipulation techniques. 

\subsection{Adversarial Threat Model}

We consider an inference-time evasion threat in which the attacker seeks to cause a manipulated image to be classified as authentic, without modifying the detector, its parameters, or its training data. For source detector \(s\), let \(p_s(x)\) denote the predicted probability of the fake class for an image $x$, and let \(\tau_s\) denote its operational decision threshold. We define the threshold-adjusted margin as
\[
m_s(x)
=
\operatorname{logit}\!\left(p_s(x)\right)
-
\operatorname{logit}\!\left(\tau_s\right).
\]
Where \(m_s(x)\geq 0\) corresponds to a fake prediction, whereas \(m_s(x)<0\) corresponds to an authentic prediction. For each deepfake image \(x_i\) and attack $a$, the attacker seeks a perturbation, $\delta$, by solving
\[
\delta_i^{*,a}
=
\arg\min_{\substack{
    \lVert\delta\rVert_\infty\leq\epsilon\\
    x_i+\delta\in[0,1]^d
}}
\mathcal{L}_{a}\!\left(m_s(x_i+\delta)\right)
\]
Here, \(\mathcal{L}_{a}\) denotes the attack-specific objective formulated with respect to the threshold-adjusted margin. We set the perturbation budget to \(\epsilon=8/255\). The resulting adversarial example is given by 
\(x_i^{\mathrm{adv},s,a}=x_i+\delta_i^{*,a}\).

We instantiate this threshold-aware evasion objective using AutoAttack (AA) \cite{autoattack} and an \(L_\infty\)-constrained CW-style margin attack with Expectation over Transformation (CW--EOT)~\cite{cw,eot}. Adversarial examples are generated independently for each source detector. For AA, we expose the threshold-adjusted two-class logits \(\widetilde{f}_s(x)=[0,m_s(x)]\), such that its constituent attacks operate relative to the source detector's operational threshold. We use a custom AA configuration comprising APGD-CE, FAB, and Square-Attack. APGD-DLR is omitted because its standard loss requires at least three class logits and is therefore not applicable to our binary surrogate logit representation.

For \(a=\mathrm{CW\text{-}EOT}\), the attack-specific objective is
\[
\mathcal{L}_{a}(x_i,\delta)
=
\mathbb{E}_{T\sim\mathcal{T}}
\left[
\max\left\{
m_s\!\left(T(x_i+\delta)\right)+\kappa,\,0
\right\}
\right].
\]
where \(\mathcal{T}\) denotes the distribution of transformations and \(\kappa\) controls the desired confidence margin. We optimize for 100 iterations with a learning rate of \(0.01\), using 10 EOT samples per
iteration. The transformations comprise random rotations of up to \(\pm3^\circ\), translations of up to \(3\%\) of the image dimensions, isotropic scaling in \([0.97,1.03]\), and additive Gaussian noise with standard deviation \(0.01\). We set \(\kappa=1\).

The attacker has white-box access to the source detector \(s\) during adversarial generation. Evaluation on \(t=s\) therefore represents the white-box setting, and transfer to any target detector \(t\neq s\) represents the black-box, where the attacker has no access to the target's parameters, gradients, or predictions during generation.

\subsection{Evaluation Metrics} 

For each attack \(a\in\mathcal \{\mathrm{AA},\mathrm{CW\text{-}EOT}\}\) and ordered source--target pair \((s,t)\), we compute attack success rate (ASR) over the manipulated images correctly classified by the target before perturbation, \( \mathcal{C}_t=\left\{i:m_t(x_i)\geq 0\right\} \). Transfer success is defined as

\[
\mathrm{ASR}^{a}_{s\rightarrow t}
=
\frac{1}{|\mathcal{C}_t|}
\sum_{i\in\mathcal{C}_t}
\mathbf{1}\!\left[
m_t\!\left(x_i^{\mathrm{adv},s,a}\right)<0
\right].
\]

Thus, \(\mathrm{ASR}^{a}_{s\rightarrow t}\) represents the proportion of manipulated images initially correctly classified by the target detector that are misclassified as authentic after the attack. By conditioning on initially correct predictions, ASR separates attack-induced failures from pre-existing detection errors, facilitating comparisons across target detectors with different clean performance, as is standard in adversarial robustness evaluation \cite{Serrano2026DUMB}.

In addition to pairwise ASR, we measure target-level vulnerability using two empirical black-box oracle protocols. The \textbf{self-excluding oracle} (SE) measures worst-case black-box vulnerability within the evaluated model pool. It considers every detector other than the target as an eligible surrogate, placing no restrictions on similarity in backbone or training data: \[ \mathcal{S}^{\mathrm{SE}}_t = \mathcal{M}\setminus\{t\}. \]

The \textbf{strict oracle} (STR) models an attacker who cannot construct a surrogate that shares the target's exact backbone or contains its complete manipulated-image training set, for example because neither is publicly available. Eligible surrogates therefore satisfy \(b_s \neq b_t\) and \(d_t \nsubseteq d_s\). Formally,
\[
\mathcal{S}^{\mathrm{STR}}_t
=
\left\{
s\in\mathcal{M}:
s\neq t,\;
b_s\neq b_t,\;
d_t\nsubseteq d_s
\right\},
\]

Note that this excludes considering models trained on the ALL subset as sources, but not targets. Importantly, it permits partial overlap between the source and target training data, reflecting
realistic settings in which a target is trained on a mixture of publicly available and private data: the attacker may reproduce the public component without having access to the target's complete training set.

For oracle setting \(q\in\{\mathrm{SE},\mathrm{STR}\}\), the oracle ASR is 
\[
\mathrm{OracleASR}^{q}_{t}
=
\frac{1}{|\mathcal{C}_t|}
\sum_{i\in\mathcal{C}_t}
\mathbf{1}\!\left[
\left(
\min_{\substack{
s\in\mathcal{S}^{q}_t\\
a\in\mathcal{A}
}}
m_t\!\left(x_i^{\mathrm{adv},s,a}\right)
\right)
<0
\right],
\]
where \(\mathcal{A}\) denotes the set of attacks available to the oracle.

\subsection{Statistical Analysis}
\label{sec:statistical_analysis}

Simple grouped averages can be misleading because source--target pairs grouped by one characteristic may also differ in several others. Therefore, we use equally weighted stratified contrasts. For each characteristic, we divide the source--target pairs into strata, with each stratum grouping pairs with the same values for the other properties relevant to that comparison. For example, when evaluating exact-backbone compatibility, each stratum fixes the pretraining regime and training data separately for the source and target. Within each eligible stratum, we compare the mean ASR of pairs that share a backbone with the mean ASR of pairs that do not. We then average these differences equally across eligible strata to estimate the average adjusted contrast for that characteristic. This common-weight construction is a special case of adjustment by subclassification \cite{cochran1968effectiveness}. We apply the same procedure to architecture family, pretraining regime, and training data. A positive contrast indicates that sharing the evaluated characteristic is associated with higher transfer. All comparisons are restricted to black-box pairs. For the training-data comparison, we exclude pairs involving \textsc{All}-trained detectors because \textsc{All} contains all four individual training subsets. For the architecture-family comparison, we exclude same-backbone pairs so that it measures sharing the broader CNN/Transformer family rather than an exact backbone.

We estimate the standard error of our estimator using a leave-one-node-out jackknife, deleting each detector as both
source and target and recomputing the complete statistic \cite{lin2020network}. We compute the jackknife variance using the conventional delete-one estimator \cite{Efron_Hastie_2016}, and use the resulting standard errors to construct approximate 95\% Wald confidence intervals and two-sided $p$-values under a standard-normal reference distribution~\cite{casella2002wald}.

In the main text, we report 13 hypothesis tests: four compatibility contrasts for each attack (shared exact backbone, architecture family, pretraining regime, and training data), four tests of whether these contrasts differ between AA and CW--EOT, and one comparison of overall mean black-box ASR between the attacks. Six additional directional contrasts are reported in the supplementary material. Since testing multiple hypotheses increases the chance of obtaining at least one false positive, we jointly adjust all 19 $p$-values using Holm's procedure, thereby controlling this probability across the full analysis \cite{holm_correction}.

\section{Experimental Results}

\subsection{Detector Performance and White-Box Attack Success}

We start by verifying that the evaluated models learned their intended detection tasks. Across the 60 detector configurations, our models attained a mean AUC of $75.3\%$ on $\mathrm{FF++}_{\mathrm{ALL}}$ and $62.6\%$ on $\mathrm{CDF}_{\mathrm{ALL}}$, with medians of $75.0\%$ and $63.0\%$ respectively. For detectors trained on a single deepfake type ($ID \in \{FS,FR,EFS,FE\}$), AUC on the corresponding held-out $\mathrm{FF++}_{\mathrm{ID}}$ domain was substantially higher, with a mean of $93.1\%$, a median of $96.0\%$, and a range of $66.0\%$--$100\%$. These results indicate that most detectors learned their designated manipulation domain, although this specialization did not always extend to the broader mixture of manipulations or to a different dataset. Complete model-level clean-performance results are reported in Table~\ref{tab:clean_performance_full} of the supplementary material.

We next verify that both attacks successfully optimize adversarial examples on their respective source detectors. Under white-box evaluation, AA achieved a mean ASR of $99.78\%$, with detector-level
values ranging from $89.66\%$ to $100.00\%$, while CW--EOT achieved a mean ASR of $99.96\%$, with values ranging from $99.00\%$ to $100.00\%$. Thus, low black-box transfer cannot be attributed to a general failure to construct successful adversarial examples on the source detectors.

\subsection{Black-Box Transfer and Source--Target Compatibility}

\begin{figure*}[t]
    \centering
    \includegraphics[width=\textwidth]
        {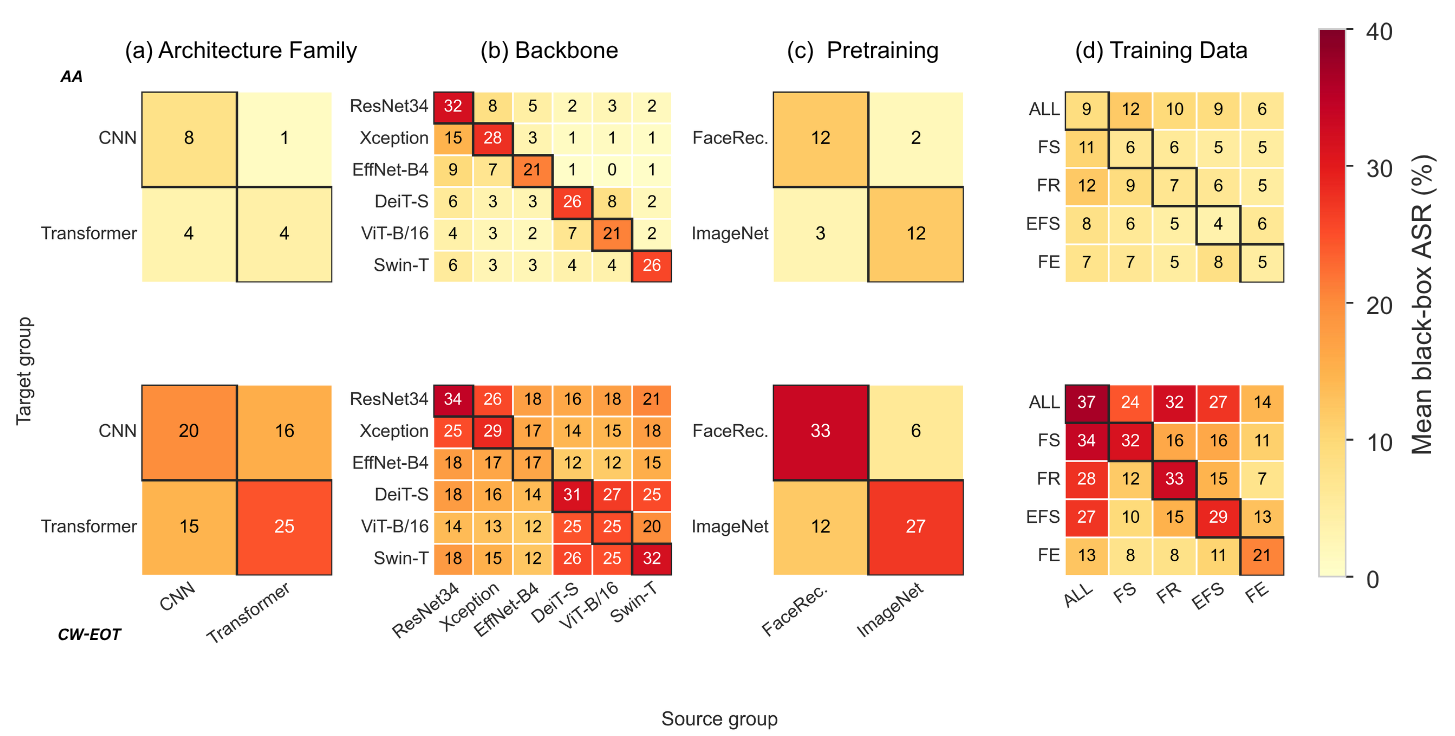}
    \caption{Mean black-box Attack Success Rate (\%) aggregated by source group
    (columns) and target group (rows) for AA (top) and CW--EOT (bottom):
    (a) architecture family, (b) backbone, (c) pretraining, and
    (d) training data. Panel (a) additionally excludes same-backbone
    pairs. Black outlines mark matched groups. White-box pairs are excluded. Columns denote source groups and rows denote target groups.}
    \label{fig:transfer_matrices_aggregated}
\end{figure*}

Having established detector functionality and white-box attack success, we examine how black-box transfer varies with the relationship between source and target detectors. Figure~\ref{fig:transfer_matrices_aggregated} reports mean black-box ASR aggregated by architecture family, exact backbone, pretraining regime, and training data. 

Under AA, the clearest differentiating effect occurs along the exact-backbone diagonal, where mean ASR ranges from $21\%$ to $32\%$, while most cross-backbone cells remain at or below $15\%$. Pretraining compatibility is also visible, where the two matched-pretraining cells both attain $12\%$ ASR, compared with $2\%$ and $3\%$ for the mismatched cells. By contrast, the architecture and training data panels exhibit weaker and less uniform diagonal structure under AA.

CW--EOT exhibits a different pattern, although we still note a strong effect in the pretraining panel. Matched Face-Recognition and ImageNet configurations attain mean ASRs of $33\%$ and $27\%$, respectively, compared with $6\%$ and $12\%$ across pretraining regimes. The main difference lies in the behavior in the backbone and training data panels. Exact-backbone transfer remains elevated under CW--EOT, but its diagonal is less uniquely dominant than under AA. A pronounced diagonal is also present among the training configurations, for which matched ALL, FS, FR, EFS, and FE cells attain $37\%$, $32\%$, $33\%$, $29\%$, and $21\%$ ASR, respectively. 

Transfer to and from \textsc{All} trained detectors and detectors trained on the FS, FR, and EFS subsets is also comparatively high, ranging from $24\%$ to $34\%$, but is lower between ALL and FE, at $13\%$--$14\%$. Because the ALL training set contains each of the four deepfake type subsets, the elevated values are consistent with training-set overlap contributing to transfer. However, the lower values involving FE indicate that overlap alone does not determine transferability. Moreover, these aggregated cells do not isolate the influence of training-set overlap from the remaining detector characteristics.

More generally, the heatmaps provide aggregated descriptive summaries of each factor. Since each cell averages over the remaining detector characteristics without matching or stratification, observed cell differences may reflect simultaneous variation in backbone, architecture family, pretraining, and training subset. Table~\ref{tab:compatibility} presents the estimates for each compatibility contrast while holding the remaining measured characteristics fixed.

\begin{table*}[t]
\centering
\caption{Matched source--target compatibility effects on black-box Attack Success Rate.
Characteristic rows compare pairs sharing versus not sharing the indicated
factor. Estimates are reported in percentage points. We highlight the \textbf{first} (bold) and \underline{second} (underlined) strongest effect per attack.}
\label{tab:compatibility}

\small
\setlength{\tabcolsep}{4pt}
\renewcommand{\arraystretch}{1.15}

\begin{tabular*}{\textwidth}{
    @{\extracolsep{\fill}}
    lrrccc
    @{}
}
\toprule
& & &
\multicolumn{2}{c}{Compatibility effect} &
Between-attack contrast \\
\cmidrule(lr){4-5}
\cmidrule(lr){6-6}

Characteristic
& $n$
& $G$
& AA
& CW-EOT
& CW-EOT${}-{}$AA \\
\midrule

Backbone
& 90
& 60
& \shortstack[c]{
    $\textbf{22.34}\ [18.88,\,25.80]$\\[-1pt]
    {\footnotesize $p_{\mathrm{Holm}}<0.0001$}
}
& \shortstack[c]{
    $13.48\ [9.71,\,17.25]$\\[-1pt]
    {\footnotesize $p_{\mathrm{Holm}}<0.0001$}
}
& \shortstack[c]{
    $-8.86\ [-11.34,\,-6.37]$\\[-1pt]
    {\footnotesize $p_{\mathrm{Holm}}<0.0001$}
}
\\

\addlinespace[3pt]

Pretraining
& 870
& 60
& \shortstack[c]{
    $\underline{9.39}\ [7.58,\,11.20]$\\[-1pt]
    {\footnotesize $p_{\mathrm{Holm}}<0.0001$}
}
& \shortstack[c]{
    $\textbf{21.25}\ [16.42,\,26.09]$\\[-1pt]
    {\footnotesize $p_{\mathrm{Holm}}<0.0001$}
}
& \shortstack[c]{
    $11.86\ [7.03,\,16.70]$\\[-1pt]
    {\footnotesize $p_{\mathrm{Holm}}<0.0001$}
}
\\

\addlinespace[3pt]

Training data\textsuperscript{a}
& 132
& 48
& \shortstack[c]{
    $3.19\ [1.68,\,4.70]$\\[-1pt]
    {\footnotesize $p_{\mathrm{Holm}}=0.0003$}
}
& \shortstack[c]{
    $\underline{19.03}\ [14.60,\,23.46]$\\[-1pt]
    {\footnotesize $p_{\mathrm{Holm}}<0.0001$}
}
& \shortstack[c]{
    $15.84\ [11.92,\,19.76]$\\[-1pt]
    {\footnotesize $p_{\mathrm{Holm}}<0.0001$}
}
\\

\addlinespace[3pt]

Architecture family\textsuperscript{b}
& 100
& 60
& \shortstack[c]{
    $3.73\ [1.70,\,5.77]$\\[-1pt]
    {\footnotesize $p_{\mathrm{Holm}}=0.0026$}
}
& \shortstack[c]{
    $7.12\ [4.35,\,9.88]$\\[-1pt]
    {\footnotesize $p_{\mathrm{Holm}}<0.0001$}
}
& \shortstack[c]{
    $3.39\ [-0.34,\,7.11]$\\[-1pt]
    {\footnotesize $p_{\mathrm{Holm}}=0.4477$}
}
\\

\midrule

Overall mean ASR
& 3540
& 60
& \multicolumn{2}{c}{\textemdash}
& \shortstack[c]{
    $12.31\ [9.41,\,15.21]$\\[-1pt]
    {\footnotesize $p_{\mathrm{Holm}}<0.0001$}
}
\\

\bottomrule
\end{tabular*}

\vspace{2pt}
\begin{minipage}{0.99\textwidth}
\footnotesize
\textsuperscript{a}\textsc{All}-trained detectors excluded.
\textsuperscript{b}Same-backbone pairs excluded.

\textit{Notes.}
Entries are estimates [95\% CI], with Holm-adjusted $p$-values below. For compatibility rows, $n$ counts matched stratum differences per attack, and in the final row, it counts source--target pairs. $G$ is the number of detector jackknife units. The final row reports the between-attack difference in overall mean ASR. CIs are unadjusted; Holm correction includes all 19 tests.
\end{minipage}

\end{table*}

Table~\ref{tab:compatibility} shows that all four compatibility contrasts are positive under both attacks and remain significant after Holm correction. Across source--target pairs, CW--EOT also achieves a mean black-box ASR \(12.31\) pp higher than AA (95\% CI \([9.41,15.21]\); \(p_{\mathrm{Holm}}<0.0001\)). More importantly, the relative importance of the compatibility factors differs between attacks.

Under AA, when source and target share the same backbone we note an average contrast in transfer (\(22.34\) pp), more than twice the increase associated with shared pretraining (\(9.39\) pp). The training data and architecture family contrasts are substantially smaller, at \(3.19\) pp and \(3.73\) pp, respectively. Under CW--EOT, the ordering changes. Shared pretraining (\(21.25\) pp) and training data (\(19.03\) pp) produce the largest contrasts, followed by exact backbone (\(13.48\) pp) and architecture family (\(7.12\) pp).

Direct between-attack comparisons confirm this shift. Relative to AA, the exact-backbone contrast is \(8.86\) pp smaller under CW--EOT, whereas the pretraining and training data contrasts are \(11.86\) pp and \(15.84\) pp larger, respectively; all three differences remain significant after Holm correction. The architecture-family contrast is also larger under CW--EOT by \(3.39\) pp, but this difference is not statistically significant (95\% CI $[-0.34,7.11]$; $p_{\mathrm{Holm}}=0.4477$).

Together, these results show that CW--EOT is not merely more transferable overall. The source--target relationships associated with transfer also differ between attacks: AA transfer is most strongly structured by exact backbone, whereas CW--EOT transfer is more strongly associated with shared pretraining and training data.

\subsection{Adversarial Vulnerability Evaluation}

\begin{figure}[!t]
    \centering
    \includegraphics[width=\columnwidth]
        {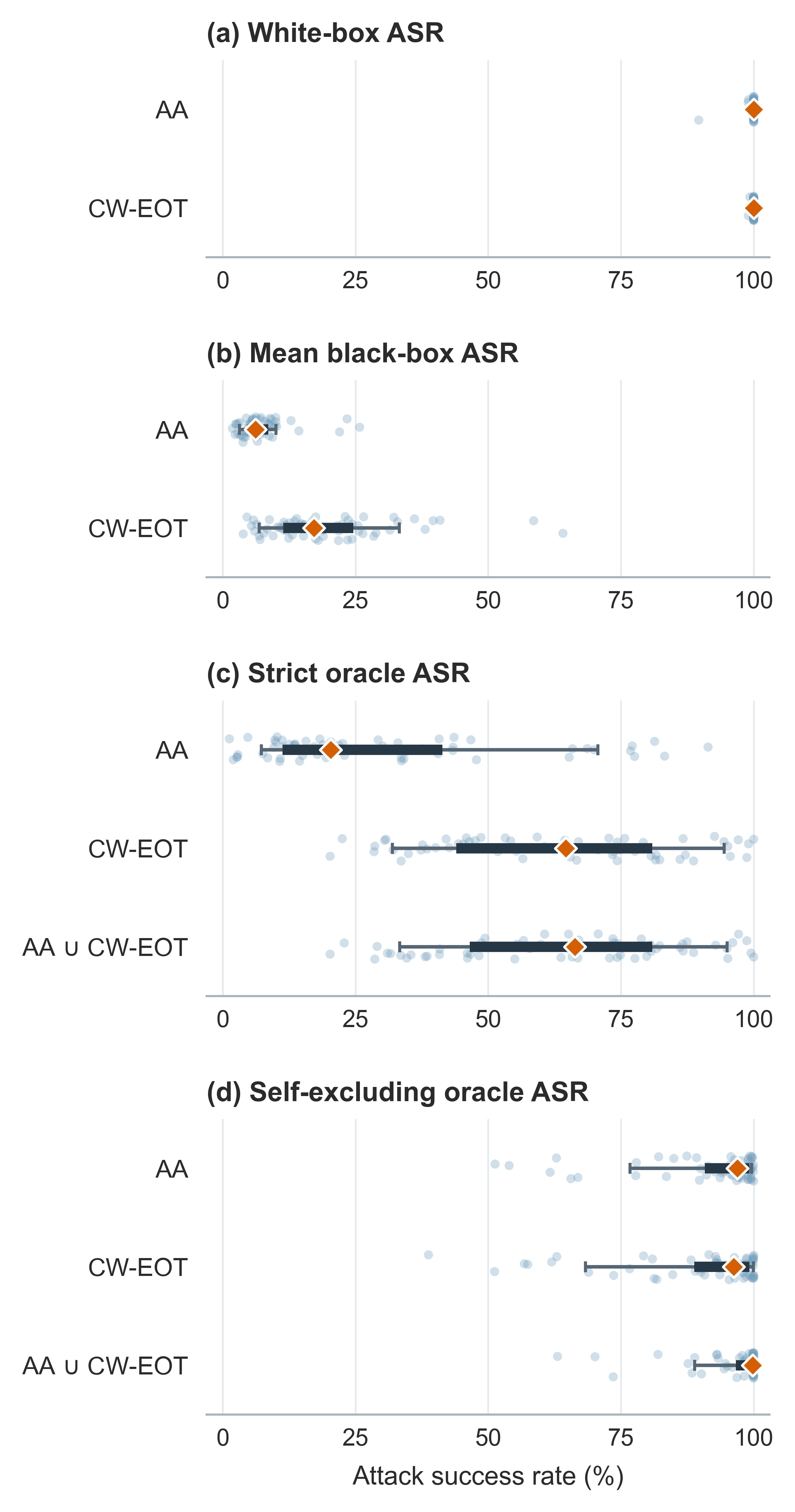}
    \caption{Per-target Attack Success Rate under white-box attacks, mean pairwise transfer, and two portfolio oracles. Points represent target detectors; diamonds, thick bars, and capped bars denote medians, interquartile ranges, and 10th--90th percentiles.}
    \label{fig:vulnerability-summary}
\end{figure}

Figure~\ref{fig:vulnerability-summary} reveals a pronounced gap between source-averaged black-box transfer and portfolio-based target vulnerability. When transfer is averaged across all non-target sources, AA achieves a mean ASR of \(7.21\%\) and a median of \(6.22\%\) across targets. Under the same evaluation, CW--EOT reaches a mean of \(19.52\%\) and a median of \(17.22\%\). Thus, a single surrogate selected without prior knowledge of its compatibility with the target appears only weakly transferable on average, although the dispersion across targets indicates substantial heterogeneity.

The result changes markedly under the self-excluding oracle. The median target-level ASR increases to \(96.94\%\) for AA and \(96.21\%\) for CW--EOT, with corresponding means of \(91.90\%\) and \(90.01\%\). Allowing candidates from either attack yields a mean ASR of \(96.37\%\) and a median of \(99.83\%\), with every target reaching at least \(63.06\%\). The similar near-saturation reached by both attacks indicates that adversarial examples generated from different sources expose complementary subsets of target inputs.

The strict-oracle results provide the more demanding comparison. Under this condition, AA attains a mean ASR of \(29.71\%\) and a median of \(20.32\%\), whereas CW--EOT retains a substantially higher mean of \(62.59\%\) and a median of \(64.60\%\). Allowing either attack produces a mean strict-oracle ASR of \(64.48\%\) and a median of \(66.34\%\). Thus, although the stricter source restrictions substantially reduce AA performance, CW--EOT continues to expose a majority of manipulated inputs for the median target.

Although these values represent portfolio upper bounds rather than expected performance without target feedback, they demonstrate that weak transfer from an individual surrogate should not be interpreted as evidence of target robustness.

\section{Discussion}
\label{sec:discussion}

\textbf{Transferability of adversarial examples strongly depends on the relationship between source and target models.} The central finding of this study is that black-box adversarial transferability is not adequately described as a property of an attack or target detector in isolation. Rather, it is heavily structured on the relationship between the source and target detectors. This structure also differs between attack types, with AA reporting a higher impact on shared backbone, while CW--EOT reports a higher impact on pretraining and training data. These patterns suggest that transfer may arise through different forms of source--target alignment. Sharing a backbone may produce similar local decision geometry or input sensitivities despite different training conditions, whereas shared data provenance may lead architecturally distinct models to learn overlapping representations or non-robust cues. AA and CW--EOT may therefore preferentially exploit different forms of alignment. Although our analysis does not directly measure gradients or representations, this interpretation implies that compatibility between two detectors is not fixed, but depends on how an attack probes their shared weaknesses.

\textbf{Shared pretraining is strongly associated with transferability.} The strong effect of shared pretraining, particularly under CW--EOT, suggests that task-specific fine-tuning does not erase the input sensitivities inherited from pretraining. These findings are consistent with prior evidence that pretraining can transmit non-robust features to fine-tuned models \cite{zhang2023imageNet} and that attacks constructed against pretrained encoders can remain effective against downstream models \cite{NEURIPS2022_084727e8}. This result makes publicly available pretrained checkpoints relevant to the black-box threat model and motivates testing whether robust pretraining can reduce downstream transferability.

\textbf{Single-source evaluation can give a false sense of robustness.} A low transfer ASR may reflect an incompatible source--target pairing, rather than evidence of robustness of the target model. Even source-averaged ASR can likewise be depressed by poorly compatible sources. Our oracle-based evaluations provide a complementary, more stringent assessment by measuring, for each input, whether any perturbation generated from an eligible surrogate pool succeeds. Our strict oracle further shows that model and data secrecy is not enough to defend against attackers that can probe the target model.

Although our controlled detector pool enabled systematic comparison of source--target factors, the conclusions remain limited to the architectures, pretraining regimes, manipulation families, datasets, and attacks evaluated here. Moreover, each detector configuration is represented by a single training run, preventing us from quantifying variability arising from stochastic training. Future work should determine whether the identified compatibility effects persist across training seeds and defended detectors, with particular attention to whether robust pretraining reduces downstream transferability.

\section{Conclusion}
\label{sec:conclusion}

Across 60 detectors and 3,540 ordered source--target pairs per attack, we find that black-box adversarial transfer is best understood as a property of the source--attack--target configuration. Transfer increased with shared backbone and training/pretraining data, but the dominant compatibility was attack-specific: exact backbone under AA, and shared pretraining and training data under CW--EOT. This places source surrogate selection as a central part of the black-box threat model. These findings have a direct consequence for robustness assessment. A low ASR from one source, or even averaging across heterogeneous source--target pairings, can conceal vulnerabilities exposed by other surrogates. Our oracle evaluations provide a complementary, more stringent assessment of target vulnerability. The strict oracle further shows that vulnerabilities extend beyond closely matched detector pipelines. Hiding model and training-data details therefore does not establish black-box robustness.

We release 240,000 adversarial images, complete pairwise transfer results, detector configurations, and evaluation code. These resources support black-box robustness evaluations that account for source–target compatibility and surrogate diversity.

{
    \small
    \bibliographystyle{ieeenat_fullname}
    \bibliography{main}
}

\clearpage
\appendix

\twocolumn[
\begin{center}
    {\LARGE\bfseries Supplementary Material}
\end{center}
\vspace{0.5em}
]

\section{Statistical Formulation and Inference}

\label{sec:supp_statistical_analysis}

This section provides support to the statistical formulation reported in the paper. We first define the contrast estimators, then describe the detector-level jackknife inference, Wald confidence intervals and \(p\)-values, and Holm adjustment used for hypothesis testing.

\subsection{Notation and Analysis Units}
\label{sec:supp_statistical_notation}

We use notation in accordance to the main paper, with \(\mathcal M\) denoting the detector bank.
For attack \(a\in\mathcal \{\mathrm{AA},\mathrm{CW\text{-}EOT}\}\), source detector \(s\), and target detector \(t\), we denote \(\mathrm{ASR}_{s\rightarrow t}^{a}\) as the attack success rate for
transfer from \(s\) to \(t\). As transfer is directional \(s\rightarrow t\) and \(t\rightarrow s\) are distinct observations.

For each attack, the eligible ordered black-box pairs are

\[
\mathcal P_a
=
\left\{
(s,t)\in\mathcal M\times\mathcal M |
s\neq t\right\}.
\]

Since the detector bank contains 60 detectors and all ordered black-box transfers are available,
\(|\mathcal P_a|=60\times59=3{,}540\) for each attack.

\subsection{Statistical Estimators}
\label{sec:supp_statistical_estimators}

Our analysis considers three distinct questions about variation in black-box transfer. First, is transfer higher when the source and target share a detector or training characteristic, after accounting for their other measured properties? Second, when transferring between two detector groups, does ASR depend on which group acts as the source and which acts as the target? Third, do AA and CW--EOT differ in their overall transferability or in how strongly source--target compatibility is associated with transfer? 

We address these questions using compatibility, directional, and between-attack contrasts, respectively. All estimators are differences in mean ASR, reported in percentage points, but each uses the comparison appropriate to its question: matched source--target configurations, reversed transfer directions, or matched observations across attacks.

\paragraph{Compatibility contrasts.}
Compatibility contrasts address whether sharing a particular characteristic is associated with higher transfer. Comparing all compatible and incompatible pairs directly could confound the characteristic of interest with the other properties of the source and target detectors. We therefore perform the comparison within matched
source--target configurations, referred to as strata.

A stratum fixes relevant properties of the source and target models. For example, when examining the effect of shared backbone, each stratum fixes the remaining properties (source pretraining, target pretraining, source training data, and target training data) to possible values. Then, within that configuration, we vary the characteristic under study and group the observations according to whether the source and target share that
characteristic. In our example of the shared backbone, we would separate the observations that share backbones between source and target from those that differ. We then compare the mean ASR of pairs from each group. 

The remaining characteristics are treated analogously. The exact-backbone and architecture-family contrasts fix source and target pretraining and training data; the pretraining contrast fixes source and target backbone
and training data; and the training-data contrast fixes source and target backbone and pretraining. Same-backbone pairs are excluded from the architecture-family contrast, while \textsc{All}-trained detectors are excluded from the training-data contrast.

More generally, let \(h\subseteq\mathcal P_a\) denote one such stratum, and let \(c(m)\) denote the value of characteristic \(c\) for detector \(m\). For an attack \(a\), the within-stratum compatibility difference is given by

\[
d_{c,h}^a
=
\underset{
\substack{
(s,t)\in h\\
c(s)=c(t)
}
}{\operatorname{mean}}
\mathrm{ASR}_{s\rightarrow t}^a
-
\underset{
\substack{
(s,t)\in h\\
c(s)\neq c(t)
}
}{\operatorname{mean}}
\mathrm{ASR}_{s\rightarrow t}^a.
\]

Only strata containing at least one compatible and one incompatible pair are retained. Let \(\mathcal H_{a,c}\) denote the set of these eligible strata for an attack \(a\) and characteristic \(c\). The compatibility estimator is given by

\[
\widehat{\Delta}^{\mathrm{comp}}_{a,c}
=
\frac{1}{|\mathcal H_{a,c}|}
\sum_{h\in\mathcal H_{a,c}}
d_{c,h}^a.
\]

\paragraph{Directional contrasts.}

Directional contrasts evaluate whether transfer between two detector groups is asymmetric. In this work we tested 3 different asymmetries:
\begin{itemize}
    \item Is transfer different when Transformer detectors act as sources and CNN detectors as targets, compared with the reverse direction?
    \item Is transfer different when detectors pretrained for face recognition act as sources and detectors pretrained on ImageNet as targets, compared with the reverse direction?
    \item Is transfer different when detectors trained on \textsc{All} act as sources and detectors trained on a single deepfake type subset as targets, compared with the reverse direction?
\end{itemize}

For comparison \(k\), let \(U_k\) and \(V_k\) denote the detector groups acting as source and target, respectively, in the first direction stated above. For attack \(a\), we calculate the ASR difference between both directions for each detector pair and average these differences:

\[
\widehat{\Delta}^{\mathrm{dir}}_{a,k}
=
\underset{
\substack{
s\in U_k,\; t\in V_k\\
(s,t),(t,s)\in\mathcal P_a
}
}{\operatorname{mean}}
\left(
\mathrm{ASR}_{s\rightarrow t}^{a}
-
\mathrm{ASR}_{t\rightarrow s}^{a}
\right).
\]

Positive values indicate greater transfer in the first direction, whereas negative values indicate greater transfer in the reverse direction.

\paragraph{Between-attack contrasts.}

We compare AA and CW--EOT in two ways: whether their compatibility
contrasts differ and whether their overall black-box ASR differs. Both
comparisons are paired over common strata or source--target pairs.

For characteristic \(c\), let
\(\mathcal H_c^\cap
=
\mathcal H_{\mathrm{CW\text{-}EOT},c}
\cap
\mathcal H_{\mathrm{AA},c}\)
denote the strata eligible under both attacks. The between-attack
difference in compatibility is

\[
\widehat{\Delta}^{\mathrm{int}}_c
=
\underset{h\in\mathcal H_c^\cap}{\operatorname{mean}}
\left(
d_{c,h}^{\mathrm{CW\text{-}EOT}}
-
d_{c,h}^{\mathrm{AA}}
\right).
\]

Positive values indicate a stronger compatibility association under
CW--EOT, whereas negative values indicate a stronger association under
AA. This directly tests whether the compatibility contrasts differ
between attacks rather than comparing their separate significance
decisions.

We also compare overall ASR over the common ordered pairs
\(\mathcal P_\cap
=
\mathcal P_{\mathrm{CW\text{-}EOT}}
\cap
\mathcal P_{\mathrm{AA}}\):

\[
\widehat{\Delta}^{\mathrm{attack}}
=
\underset{(s,t)\in\mathcal P_\cap}{\operatorname{mean}}
\left(
\mathrm{ASR}_{s\rightarrow t}^{\mathrm{CW\text{-}EOT}}
-
\mathrm{ASR}_{s\rightarrow t}^{\mathrm{AA}}
\right).
\]

Positive values indicate higher mean black-box ASR under CW--EOT.

\subsection{Statistical Inference}
\label{sec:supp_statistical_inference}

The estimators above provide point estimates of transfer differences, but their uncertainty must account for two features of our analysis. First, directed transfer observations are dependent because each detector appears
repeatedly as both a source and a target. Second, evaluating multiple hypotheses increases the probability of false-positive findings.

We therefore estimate detector-level uncertainty using a leave-one-detector-out jackknife, construct approximate Wald confidence intervals and two-sided \(p\)-values, and adjust the resulting \(p\)-values jointly using Holm's procedure.

\paragraph{Delete-one-detector jackknife.}

We use the detector, rather than the individual source--target pair, as the jackknife deletion unit. For each detector \(g\), we remove every observation in which it appears as either source or target and recompute the complete estimator. This procedure is analogous to leave-one-node-out methods for network data \cite{lin2020network}, while uncertainty is estimated using the conventional delete-one jackknife variance formula
\cite{Efron_Hastie_2016}.

Considering \(\widehat{\Delta}\) as any estimator defined above, let \(\widehat{\Delta}_{(-g)}\) denote its value after deleting detector \(g\). For compatibility-based estimators, the eligible strata and their
within-stratum differences are reconstructed after every deletion. Let \(G\) denote the number of detector deletion units. The mean of the leave-one-detector-out estimates is given by

\[
\overline{\Delta}_{(-\cdot)}
=
\frac{1}{G}
\sum_{g=1}^{G}
\widehat{\Delta}_{(-g)}.
\]

The jackknife standard error is given by

\[
\widehat{\operatorname{SE}}_{\mathrm{JK}}
(\widehat{\Delta})
=
\sqrt{
\frac{G-1}{G}
\sum_{g=1}^{G}
\left(
\widehat{\Delta}_{(-g)}
-
\overline{\Delta}_{(-\cdot)}
\right)^2
}.
\]

The estimator computed using all eligible detectors remains the reported point estimate. For most comparisons, \(G=60\). Because \textsc{All}-trained detectors do not enter the training-data compatibility contrasts, these
contrasts and their corresponding between-attack contrast use \(G=48\).

\paragraph{Wald confidence intervals and \(p\)-values.}

For any estimator \(\widehat{\Delta}\) defined above, we use its jackknife standard error and a standard-normal reference distribution to compute the approximate Wald statistic, unadjusted two-sided \(p\)-value, and unadjusted
95\% confidence interval:

\[
\begin{aligned}
z
&=
\frac{\widehat{\Delta}}
{\widehat{\operatorname{SE}}_{\mathrm{JK}}
(\widehat{\Delta})},\\
p
&=
2\Phi(-|z|),\\
\mathrm{CI}_{95\%}
&=
\widehat{\Delta}
\pm
1.96\,
\widehat{\operatorname{SE}}_{\mathrm{JK}}
(\widehat{\Delta}),
\end{aligned}
\]

where \(\Phi\) denotes the standard-normal cumulative distribution function. The \(p\)-values are subsequently adjusted using Holm's procedure, while the confidence intervals remain unadjusted.

\paragraph{Holm adjustment.}

To control the family-wise error rate across multiple hypotheses, we adjust all 19 unadjusted \(p\)-values jointly using Holm's step-down procedure \cite{holm_correction}. These comprise eight attack-specific compatibility tests, six directional tests, four between-attack compatibility tests, and one
comparison of overall ASR.

For our \(m=19\) hypotheses, let \(p_{(1)}\leq\cdots\leq p_{(m)}\) denote the ordered unadjusted \(p\)-values. The Holm-adjusted value at position \(i\) is given by

\[
p_{\mathrm{Holm (i)}}
=
\min
\left\{
1,\,
\max_{1\leq r\leq i}
\left[
(m-r+1)p_{(r)}
\right]
\right\}.
\]

The adjusted values are then returned to their original hypotheses, and
statistical significance is assessed using
\(p_{\mathrm{Holm}}<0.05\). Only the \(p\)-values are adjusted; the reported
confidence intervals remain pointwise 95\% intervals (not adjusted for multiplicity) and should not be
interpreted as simultaneous confidence intervals.

\section{Detailed Clean Performance}

Table~\ref{tab:clean_performance_full} reports clean performance for all 60 detectors. We evaluate each detector on the complete mixture of FaceForensics++ manipulations ($\mathrm{FF++}_{\mathrm{ALL}}$), the held-out FaceForensics++ subset corresponding to its DF40 training subset ($\mathrm{FF++}_{\mathrm{ID}}$), and the complete Celeb-DF test set ($\mathrm{CDF}_{\mathrm{ALL}}$). For \textsc{All}-trained detectors, the corresponding evaluation includes all FaceForensics++ manipulations and therefore coincides with $\mathrm{FF++}_{\mathrm{ALL}}$; this duplicate value is omitted.
ImageNet-pretrained detectors obtained higher mean AUC on $\mathrm{FF++}{\mathrm{ALL}}$ than face-recognition-pretrained detectors ($81.7\%$ versus $68.8\%$). Conversely, face-recognition pretraining produced moderately higher mean AUC on $\mathrm{CDF}{\mathrm{ALL}}$ ($64.8\%$ versus $60.3\%$).

\section{Additional Statistical Results}

Table~\ref{tab:additional_statistical_tests} reports the six prespecified directional contrasts not presented in the main results table. None of the tests provide evidence of a consistent transfer asymmetry after Holm adjustment.

Under AA, transfer from Transformer to CNN detectors was estimated to be 2.30 percentage points lower than in the reverse direction (95\% CI $[-4.48,-0.11]$). Although this pointwise confidence interval excludes zero, the contrast is not significant after Holm adjustment ($p_{\mathrm{Holm}}=0.2769$). The remaining AA contrasts are smaller and uncertain: detectors pretrained for face recognition do not transfer significantly better to ImageNet-pretrained detectors than in the reverse direction, and \textsc{All}-trained detectors do not exhibit a consistent advantage over detectors trained on a single subset when used as sources.

No directional contrast is significant under CW--EOT either. The largest estimate favors transfer from face-recognition-pretrained sources to ImageNet-pretrained targets by $5.96$ percentage points,
but its confidence interval includes zero (95\% CI $[-1.25,13.17]$; $p_{\mathrm{Holm}}=0.5254$). The Transformer/CNN and \textsc{All}-versus-single-subset contrasts are both small relative to their uncertainty.

These findings distinguish compatibility from directionality. Sharing a backbone, pretraining regime, or training data can increase transfer between source and target detectors, but this does not imply that one architecture, pretraining group, or training subset is consistently the stronger source. The principal transfer structure is therefore associated with source--target compatibility rather than a universal ordering between detector groups.

\section{Sensitivity Analysis}
\label{sec:supp_sensitivity}

\begin{figure*}[p]
    \centering
    \includegraphics[
        height=0.82\textheight,
        keepaspectratio
    ]{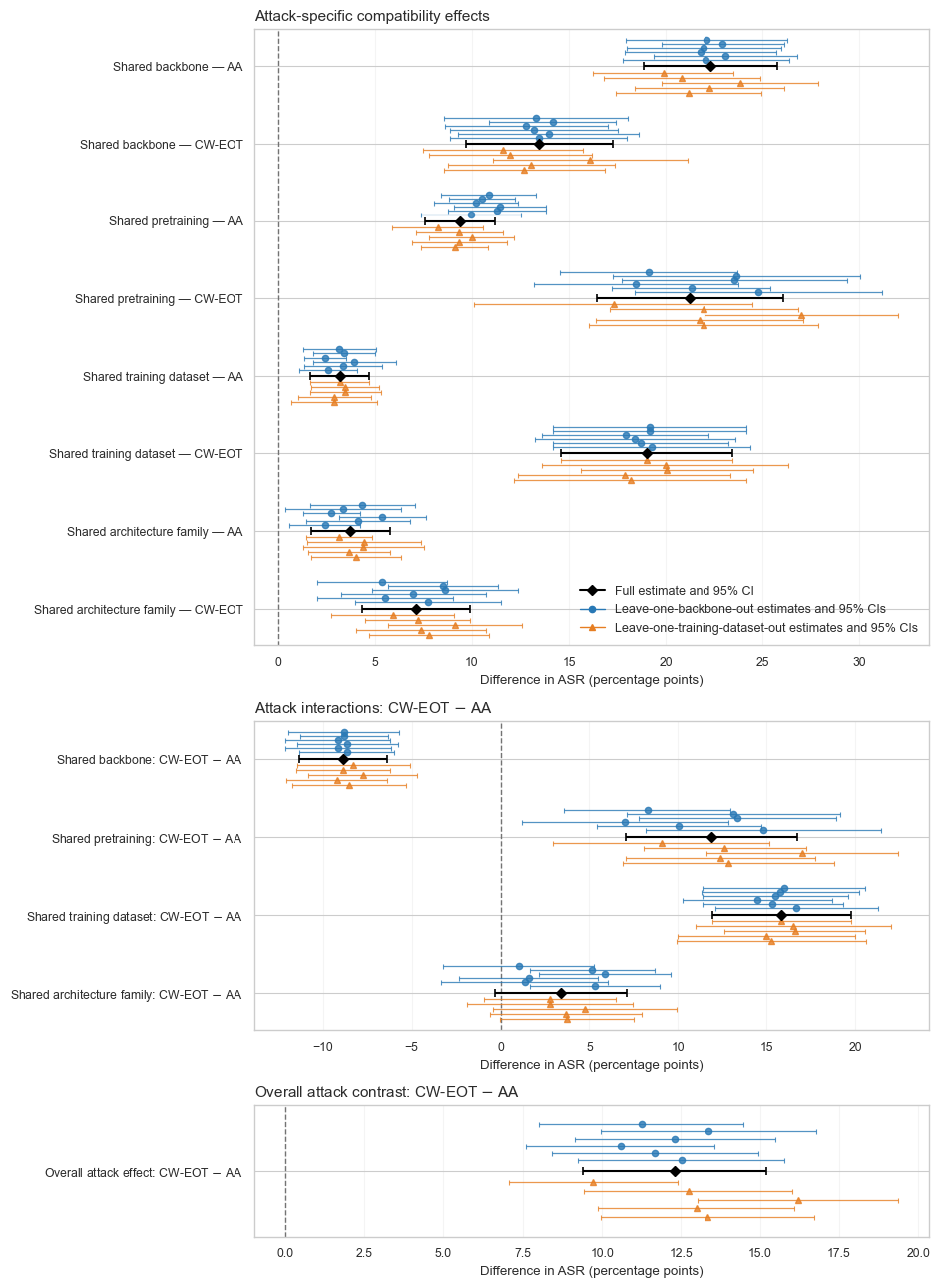}
    \caption{Sensitivity of the principal statistical estimates to detector-group omission.
    Black diamonds show estimates from the complete detector bank. Blue circles and orange triangles show estimates obtained after omitting all detectors using one backbone or trained on one DF40 subset at a time, respectively.
    Horizontal bars denote pointwise 95\% leave-one-detector-out jackknife confidence intervals recomputed using the corresponding reduced detector bank. 
    The upper panel reports attack-specific compatibility contrasts, the middle panel reports their CW--EOT-minus-AA differences, and the lower panel reports the overall paired contrast between attacks. The omission estimates are descriptive diagnostics and do not constitute additional hypothesis tests.}
    \label{fig:loo_sensitivity}
\end{figure*}

We further examined whether the principal findings were disproportionately determined by a particular backbone or manipulation-training dataset. For each sensitivity replicate, we removed all detectors associated with one backbone or one training dataset and recomputed the complete estimator on the reduced detector bank. Figure \ref{fig:loo_sensitivity} compares these group-omission estimates with the corresponding estimates obtained from the complete detector bank.

The signs and qualitative ordering of the principal compatibility effects remain stable across the omissions. Exact-backbone compatibility remains the dominant effect under AA, whereas shared pretraining and shared training data remain the strongest effects under CW--EOT. Similarly, every group-omission replicate preserves the negative interaction for exact-backbone compatibility and the positive interactions for pretraining and training data compatibility. The overall CW--EOT versus AA contrast also remains positive after every omission.

The smaller AA training dataset and architecture-family effects exhibit greater relative variation, while the architecture-family interaction remains the least precisely estimated interaction. This is consistent with the main analysis, in which the latter interaction does not differ significantly from zero. Nevertheless, no individual backbone or training dataset reverses the central pattern of attack-specific compatibility effects.

\section{Detailed Transfer Results}

\begin{figure*}[t]
    \centering
    \includegraphics[width=\textwidth]
        {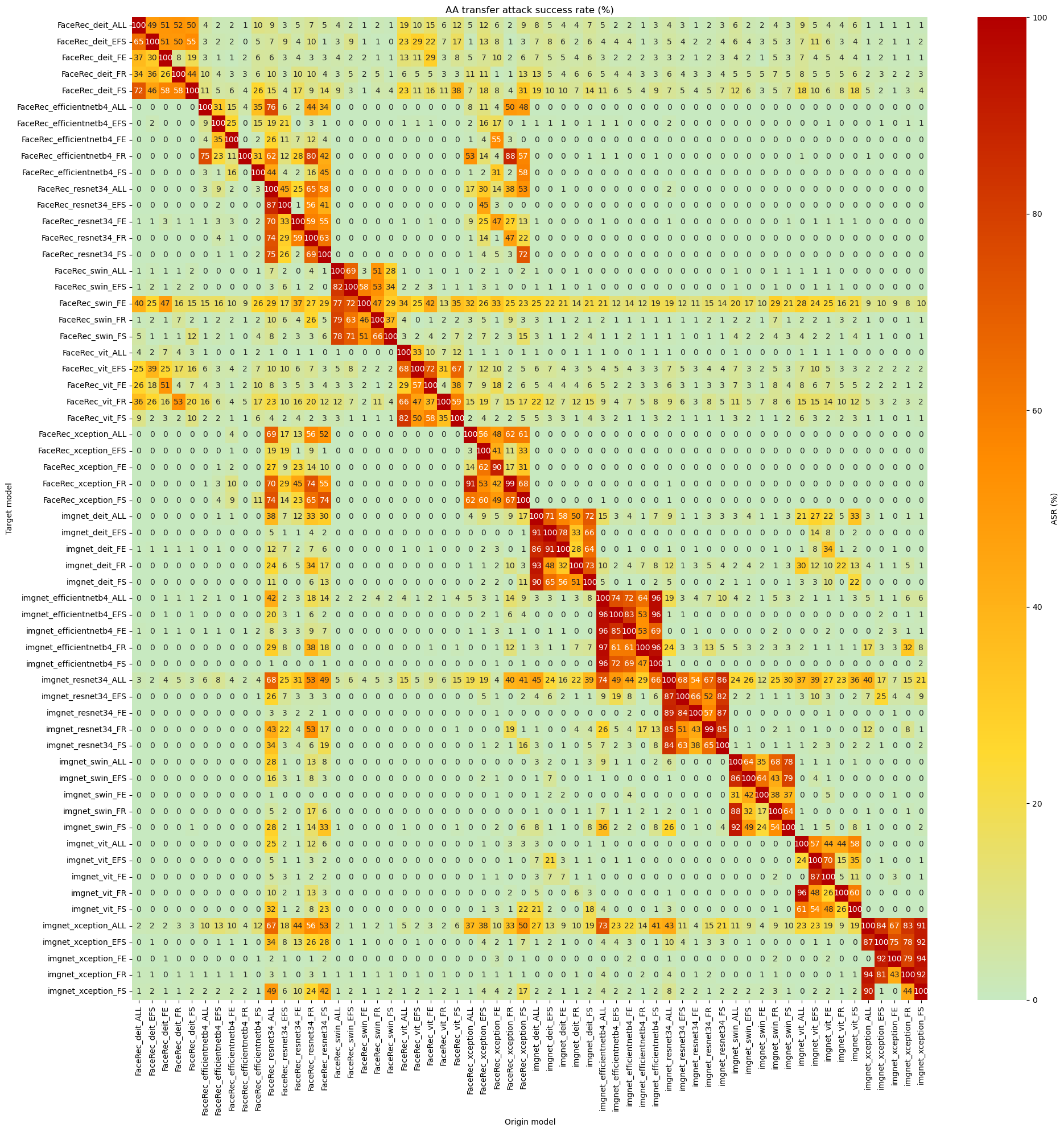}
    \caption{Full pairwise transfer matrix for AA. Each cell reports the attack
    success rate (ASR, \%) when the column detector is used as the source and the
    row detector as the target. Diagonal entries correspond to white-box
    evaluation, whereas off-diagonal entries correspond to black-box transfer.}
    \label{fig:aa_transfer_matrix}
\end{figure*}

\begin{figure*}[t]
    \centering
    \includegraphics[width=\textwidth]
        {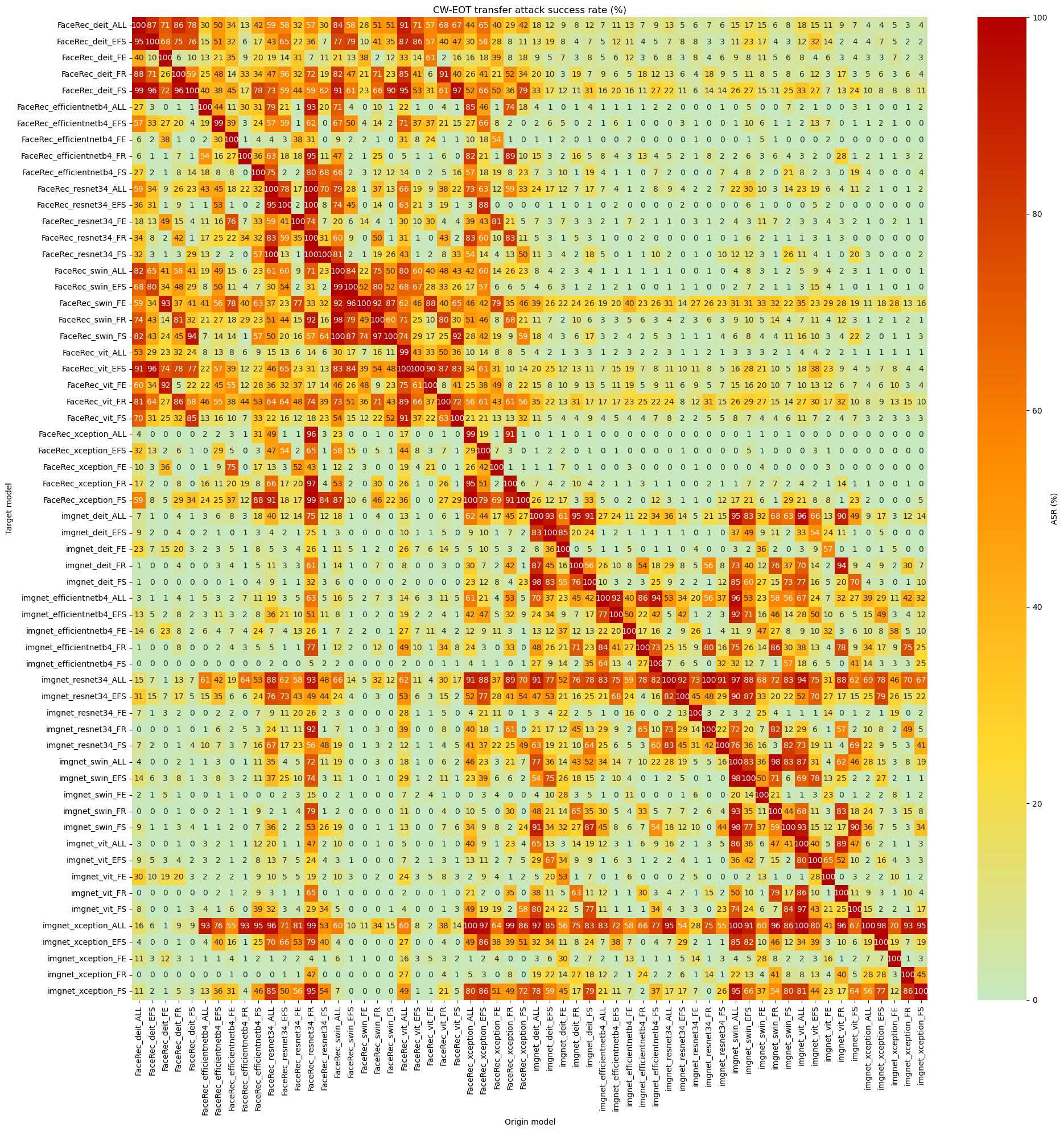}
    \caption{Full pairwise transfer matrix for CW--EOT. Each cell reports the
    attack success rate (ASR, \%) when the column detector is used as the source
    and the row detector as the target. Diagonal entries correspond to white-box
    evaluation, whereas off-diagonal entries correspond to black-box transfer.}
    \label{fig:cw_transfer_matrix}
\end{figure*}

Figures \ref{fig:aa_transfer_matrix} and \ref{fig:cw_transfer_matrix} present the complete model-level transfer matrices for AA and CW--EOT, respectively. The matrices reveal substantial variation across source--target combinations that is concealed by aggregate transfer estimates. AA transfer is comparatively sparse and concentrated among particular compatible detector configurations, whereas CW--EOT generally produces broader and stronger transfer. Nevertheless, individual matrix entries may reflect simultaneous differences in backbone, architecture family, pretraining, and training data. Consequently, the effects of these characteristics are assessed using the stratified contrasts reported in Table \ref{tab:compatibility}, rather than
through isolated comparisons between matrix entries.

\section{Experimental and Reproducibility Details}
\label{sec:supp_experimental_details}

\paragraph{Face Recognition pretraining.}
The face recognition pretraining was obtained by training each backbone from random initialization on all four BUPT-BalancedFace ethnicity partitions \cite{balancedface}. Images were resized to \(224\times224\), and the backbone output was projected to a 512-dimensional, \(\ell_2\)-normalized embedding. Identity classification used ElasticArcFace+ \cite{elasticface} with margin \(m=0.3\), sampling standard deviation \(0.05\), and scale 64 for CNNs or 32 for Transformers. Training augmentation comprised horizontal flipping (\(p=0.5\)), color jitter (\(p=0.3\)), and random affine transformations with rotations up to \(5^\circ\), translations up to \(3\%\), and scaling in \([0.97,1.03]\) (\(p=0.3\)).

All backbones were trained for 30 epochs with batch size 256. The CNNs used SGD with momentum \(0.9\), weight decay \(5\times10^{-4}\), and effective learning rates of \(10^{-2}\), \(5\times10^{-3}\), and \(2.5\times10^{-3}\) for ResNet-34, Xception, and EfficientNet-B4, respectively. The Transformers used AdamW with weight decay \(0.05\), three warm-up epochs, cosine decay, and effective learning rates of \(10^{-4}\) for DeiT-S and Swin-T and \(5\times10^{-5}\) for ViT-B/16. The checkpoint with the highest RFW \cite{rfw} validation AUC was retained, and only its backbone weights were transferred to detector training.

\paragraph{Detector training.}
The \textsc{FS}, \textsc{FR}, \textsc{EFS}, and \textsc{FE} configurations used their respective DF40 manipulation subsets, while \textsc{All} used their union. All models used the same real-image partition and a common validation mixture containing the four manipulation subsets. One indexed frame was used per video.

Training augmentation consisted of horizontal flipping with probability \(0.5\), and Gaussian blur with probability \(0.5\) (kernel sizes from 3 to 7). With probability \(0.5\), one of three color augmentations was selected with equal probability: random brightness--contrast adjustment, PCA-based color perturbation, or hue--saturation adjustment. JPEG compression with quality sampled from 40 to 100 was applied with probability \(0.5\). Inputs were \(224\times224\), except for ImageNet-initialized Xception, which used \(256\times256\). ImageNet initialization used ImageNet normalization except for Xception, whereas face-recognition initialization used mean and standard deviation \((0.5,0.5,0.5)\).

\paragraph{Hyperparameter selection and optimization.}
For the face recognition pretrained detectors, we conducted architecture-specific adaptive Bayesian searches using the \textsc{All} configuration and validation AUC as the objective. The search considered separate backbone and classification-head learning rates, weight decay, and, for Transformers, dropout, attention dropout, stochastic depth, and warm-up. Additional trials were allocated when initial configurations failed to converge reliably. After selection, the architecture-specific settings were fixed across all five training-data conditions.

The final settings are reported in Table~\ref{tab:experimental-details}, with unrounded values provided in the released YAML files. All models were fine-tuned end-to-end using two-class cross-entropy and cosine learning-rate annealing to \(10^{-6}\). We train the models for 51 epochs, with a fixed seed (42). Validation was performed after every epoch, and the checkpoint with the highest validation AUC was retained.

\paragraph{Threshold selection and reproducibility.}
For each retained detector, 1,001 thresholds uniformly spaced over \([0,1]\) were evaluated on the validation split. The threshold maximizing balanced accuracy was selected, with ties resolved in favor of the value closest to \(0.5\), and was then fixed for clean and adversarial evaluation.


\begin{table*}[!t]
\centering
\caption{Clean AUC across the 60 detectors, organized by pretraining
regime, backbone, and DF40 training subset. Entries report
\(\mathrm{FF\!+\!+}_{\mathrm{ALL}}/
\mathrm{FF\!+\!+}_{\mathrm{ID}}/
\mathrm{CDF}_{\mathrm{ALL}}\) AUC in percent, rounded to the nearest
percentage point. For \textsc{All}-trained detectors, the duplicated
in-domain value is omitted (\text{--}).}

\label{tab:clean_performance_full}

\small
\setlength{\tabcolsep}{2.5pt}
\renewcommand{\arraystretch}{0.98}

\begin{tabular*}{\textwidth}{
    @{\extracolsep{\fill}}
    ll@{\hspace{7pt}}ccccc
    @{}
}
\toprule
& & \multicolumn{5}{c}{\textbf{Detector training subset}} \\
\cmidrule(lr){3-7}
\textbf{Pretraining} & \textbf{Backbone}
& \textsc{All} & FS & FR & EFS & FE \\
\midrule

\multirow{6}{*}{ImageNet}
& ResNet34
& $93/\text{--}/60$
& $83/92/56$
& $85/98/48$
& $76/99/75$
& $75/100/65$ \\

& Xception
& $96/\text{--}/57$
& $88/95/60$
& $88/97/40$
& $83/100/75$
& $76/100/66$ \\

& EffNet-B4
& $94/\text{--}/69$
& $81/93/67$
& $83/99/40$
& $75/99/76$
& $75/100/61$ \\

\addlinespace[1pt]

& DeiT-S
& $86/\text{--}/70$
& $74/88/49$
& $74/94/50$
& $72/98/76$
& $68/100/69$ \\

& ViT-B/16
& $90/\text{--}/73$
& $84/95/53$
& $84/99/48$
& $76/99/76$
& $67/100/60$ \\

& Swin-T
& $96/\text{--}/53$
& $91/98/45$
& $91/100/38$
& $76/98/76$
& $71/100/59$ \\

\midrule

\multirow{6}{*}{FaceRec.}
& ResNet34
& $84/\text{--}/73$
& $62/84/62$
& $69/85/63$
& $75/94/63$
& $66/100/63$ \\

& Xception
& $90/\text{--}/68$
& $75/87/61$
& $78/90/59$
& $73/94/62$
& $68/100/66$ \\

& EffNet-B4
& $84/\text{--}/63$
& $74/82/65$
& $74/94/55$
& $67/90/55$
& $67/100/62$ \\

\addlinespace[1pt]

& DeiT-S
& $75/\text{--}/78$
& $59/72/62$
& $63/78/69$
& $67/94/68$
& $61/100/62$ \\

& ViT-B/16
& $71/\text{--}/68$
& $51/66/61$
& $64/77/64$
& $65/93/73$
& $63/100/60$ \\

& Swin-T
& $74/\text{--}/77$
& $59/76/67$
& $66/80/66$
& $66/93/73$
& $55/98/56$ \\

\bottomrule
\end{tabular*}
\end{table*}

\begin{table*}[t]
\centering
\caption{ Directional contrasts in black-box ASR. Each row reports the mean paired difference between transfer in the first displayed direction and transfer in the reverse direction, separately for AA and CW--EOT. Estimates are reported in percentage points}
\label{tab:additional_statistical_tests}

\small
\setlength{\tabcolsep}{5pt}
\renewcommand{\arraystretch}{1.15}

\begin{tabular*}{\textwidth}{
    @{\extracolsep{\fill}}
    lrrcc
    @{}
}
\toprule
& & &
\multicolumn{2}{c}{Directional contrast} \\
\cmidrule(lr){4-5}

Contrast
& $n$
& $G$
& AA
& CW-EOT \\
\midrule

Transformer${}\rightarrow{}$CNN
vs.\ CNN${}\rightarrow{}$Transformer
& 900
& 60
& \shortstack[c]{
    $-2.30\ [-4.48,\,-0.11]$\\[-1pt]
    {\footnotesize $p_{\mathrm{Holm}}=0.2769$}
}
& \shortstack[c]{
    $0.89\ [-7.32,\,9.10]$\\[-1pt]
    {\footnotesize $p_{\mathrm{Holm}}=1.0000$}
}
\\

\addlinespace[3pt]

FaceRec.${}\rightarrow{}$ImageNet
vs.\ ImageNet${}\rightarrow{}$FaceRec.
& 900
& 60
& \shortstack[c]{
    $0.94\ [-2.06,\,3.94]$\\[-1pt]
    {\footnotesize $p_{\mathrm{Holm}}=1.0000$}
}
& \shortstack[c]{
    $5.96\ [-1.25,\,13.17]$\\[-1pt]
    {\footnotesize $p_{\mathrm{Holm}}=0.5254$}
}
\\

\addlinespace[3pt]

\textsc{All}${}\rightarrow{}$single subset
vs.\ single subset${}\rightarrow{}$\textsc{All}
& 576
& 60
& \shortstack[c]{
    $0.29\ [-5.17,\,5.76]$\\[-1pt]
    {\footnotesize $p_{\mathrm{Holm}}=1.0000$}
}
& \shortstack[c]{
    $0.81\ [-14.92,\,16.53]$\\[-1pt]
    {\footnotesize $p_{\mathrm{Holm}}=1.0000$}
}
\\

\bottomrule
\end{tabular*}

\vspace{2pt}
\begin{minipage}{0.99\textwidth}
\footnotesize
\textit{Notes.}
Entries report the estimate [95\% leave-one-detector-out jackknife CI], with the Holm-adjusted $p$-value below. Positive estimates indicate higher mean black-box ASR in the first displayed direction, whereas negative estimates favor the reverse direction. Here, $n$ is the number of paired bidirectional detector comparisons contributing to each attack-specific contrast, and $G$ is the number of detector deletion units. Confidence intervals are unadjusted for multiplicity, whereas $p$-values are Holm-adjusted jointly across the 19 prespecified tests. White-box pairs are excluded.
\end{minipage}

\end{table*}

\begin{table*}[t]
\centering
\caption{Final detector-training configurations. $\eta_B$ and $\eta_H$ denote the backbone and classification-head learning rates, respectively. WD denotes weight decay. WU denotes the number of warm-up epochs.}
\label{tab:experimental-details}
\scriptsize
\setlength{\tabcolsep}{4pt}
\renewcommand{\arraystretch}{1.08}
\begin{tabular}{llrrlrrl}
\toprule
Initialization & Backbone & Resolution & Batch & Optimizer & $\eta_B$ & $\eta_H$ & WD / WU \\
\midrule
\multicolumn{8}{l}{\textit{ImageNet}} \\
\addlinespace[1pt]
& ResNet-34        & 224 & 128 & Adam  & $1.00\times 10^{-4}$ & $1.00\times 10^{-4}$ & $1.00\times 10^{-3}$ / 0 \\
& Xception         & 256 & 128 & Adam  & $1.00\times 10^{-4}$ & $1.00\times 10^{-4}$ & $1.00\times 10^{-3}$ / 0 \\
& EfficientNet-B4  & 224 & 128 & Adam  & $1.00\times 10^{-4}$ & $1.00\times 10^{-4}$ & $1.00\times 10^{-3}$ / 0 \\
& DeiT-S           & 224 & 128 & Adam  & $1.00\times 10^{-5}$ & $1.00\times 10^{-4}$ & $1.00\times 10^{-3}$ / 0 \\
& ViT-B/16         & 224 & 128 & AdamW & $2.00\times 10^{-5}$ & $3.00\times 10^{-4}$ & $1.00\times 10^{-2}$ / 5 \\
& Swin-T           & 224 & 64  & Adam  & $1.00\times 10^{-5}$ & $1.00\times 10^{-4}$ & $1.00\times 10^{-3}$ / 0 \\
\midrule
\multicolumn{8}{l}{\textit{Face recognition}} \\
\addlinespace[1pt]
& ResNet-34        & 224 & 128 & AdamW & $4.91\times 10^{-5}$ & $7.50\times 10^{-5}$ & $1.79\times 10^{-3}$ / 0 \\
& Xception         & 224 & 128 & AdamW & $1.47\times 10^{-4}$ & $1.49\times 10^{-4}$ & $1.17\times 10^{-3}$ / 0 \\
& EfficientNet-B4  & 224 & 128 & AdamW & $1.86\times 10^{-4}$ & $1.34\times 10^{-5}$ & $8.91\times 10^{-4}$ / 0 \\
& DeiT-S           & 224 & 128 & AdamW & $6.80\times 10^{-5}$ & $5.98\times 10^{-4}$ & $3.73\times 10^{-6}$ / 5 \\
& ViT-B/16         & 224 & 128 & AdamW & $6.00\times 10^{-5}$ & $6.50\times 10^{-4}$ & $4.00\times 10^{-4}$ / 5 \\
& Swin-T           & 224 & 64  & AdamW & $4.50\times 10^{-5}$ & $7.82\times 10^{-5}$ & $3.97\times 10^{-3}$ / 5 \\
\bottomrule
\end{tabular}
\end{table*}

\end{document}